\documentclass[runningheads]{llncs}
\usepackage{times}  
\usepackage{helvet}  
\usepackage{courier}  
\usepackage{url}  
\usepackage{graphicx}  
\usepackage{times}
\usepackage{xcolor}
\usepackage{soul}
\usepackage[utf8]{inputenc}
\usepackage[small]{caption}
\usepackage{comment}
\usepackage{color}

\usepackage{epsfig}
\usepackage{graphicx}
\usepackage{amsmath}
\usepackage{amssymb}
\usepackage{booktabs}       
\usepackage{wrapfig}

\usepackage{url}
\usepackage{multirow}
\usepackage{enumitem}
\usepackage{moresize}
\usepackage{epstopdf}
\usepackage{array}

\usepackage[ruled,vlined,linesnumbered]{algorithm2e}

\usepackage{wrapfig}
\usepackage{subfigure}
\usepackage[hang,flushmargin]{footmisc} 

\let\OLDthebibliography\thebibliography
\renewcommand\thebibliography[1]{
  \OLDthebibliography{#1}
  \setlength{\parskip}{1.6pt}
  \setlength{\itemsep}{0pt plus 0.3ex}
}

\usepackage{xspace}
\makeatletter
\DeclareRobustCommand\onedot{\futurelet\@let@token\@onedot}
\def\@onedot{\ifx\@let@token.\else.\null\fi\xspace}
\def\eg{\emph{e.g}\onedot} 
\def\ie{\emph{i.e}\onedot} 
 
\def\etc{\emph{etc}\onedot}

\makeatother

\definecolor{darkgreen}{rgb}{0,0.694,0.298}
\definecolor{purple}{rgb}{0.4,0.176,0.569}

\begin{document}

\clubpenalty = 10000
\widowpenalty = 10000
\displaywidowpenalty = 10000

\title{Metamorphic Relation Based Adversarial Attacks on Differentiable Neural Computer}

\author{Alvin Chan\inst{1} 
\and Lei Ma\inst{2,1} \and
Felix Juefei-Xu\inst{3}\and
Xiaofei Xie\inst{1}\\
Yang Liu\inst{1}\and
Yew Soon Ong\inst{1}}


\institute{Nanyang Technological University \and Harbin Institute of Technology \and
Carnegie Mellon University}

\maketitle

\begin{abstract}
Deep neural networks (DNN), while becoming the driving force of many novel technology and achieving tremendous success in many cutting edge applications, are still vulnerable to adversarial attacks. Differentiable neural computer (DNC) is a novel computing machine with DNN as its central controller operating on an external memory module for data processing. The unique architecture of DNC contributes to its state-of-the-art performance in tasks which requires the ability to represent variables and data structure as well as to store data over long timescales. However, there still lacks a comprehensive study on how adversarial examples affect DNC in terms of robustness. In this paper, we propose metamorphic relation based adversarial techniques for a range of tasks described in the natural processing language domain. We show that the near-perfect performance of the DNC in bAbI logical question answering tasks can be degraded by adversarially injected sentences. We further perform in-depth study on the role of DNC’s memory size in its robustness and analyze the potential reason causing why DNC fails. Our study demonstrates the current challenges and potential opportunities towards constructing more robust DNCs.


\end{abstract}

\section{Introduction}
Over the past decades, deep neural networks (DNN) experienced unprecedented rapid development in company with the data explosion, and achieves impressive success in matching human intelligence in many applications, such as IBM's Watson \cite{kelly2013smart}, DeepMind's Atari \cite{mnih2015human} and AlphaGo \cite{alphago}.
However, extensive studies reveal that DNN is vulnerable to adversarial attacks \cite{szegedy2013intriguing,goodfellow2014explaining,Papernot2017,pei2017deepxplore,tian2017deeptest,madry2017towards}, where imperceptible perturbations to an input can cause DNN to misclassify images with high confidence. This could impede DNN's application especially in safety critical scenarios, and cause losses and even severe tragedies if flawed DNNs are deployed to safety-critical systems (\eg, the recent Tesla autonomous driving accident \cite{tesla}). Such DNN vulnerabilities have since been also studied in other domains such as reinforcement learning \cite{behzadan2017vulnerability,huang2017adversarial}, speech recognition \cite{adv_speech} and some natural language processing (NLP) tasks \cite{Alzantot,advcompre,papernot2016crafting,kuleshov2018adversarial}.
\begin{figure}[t]
\centering
\includegraphics[width=\columnwidth]{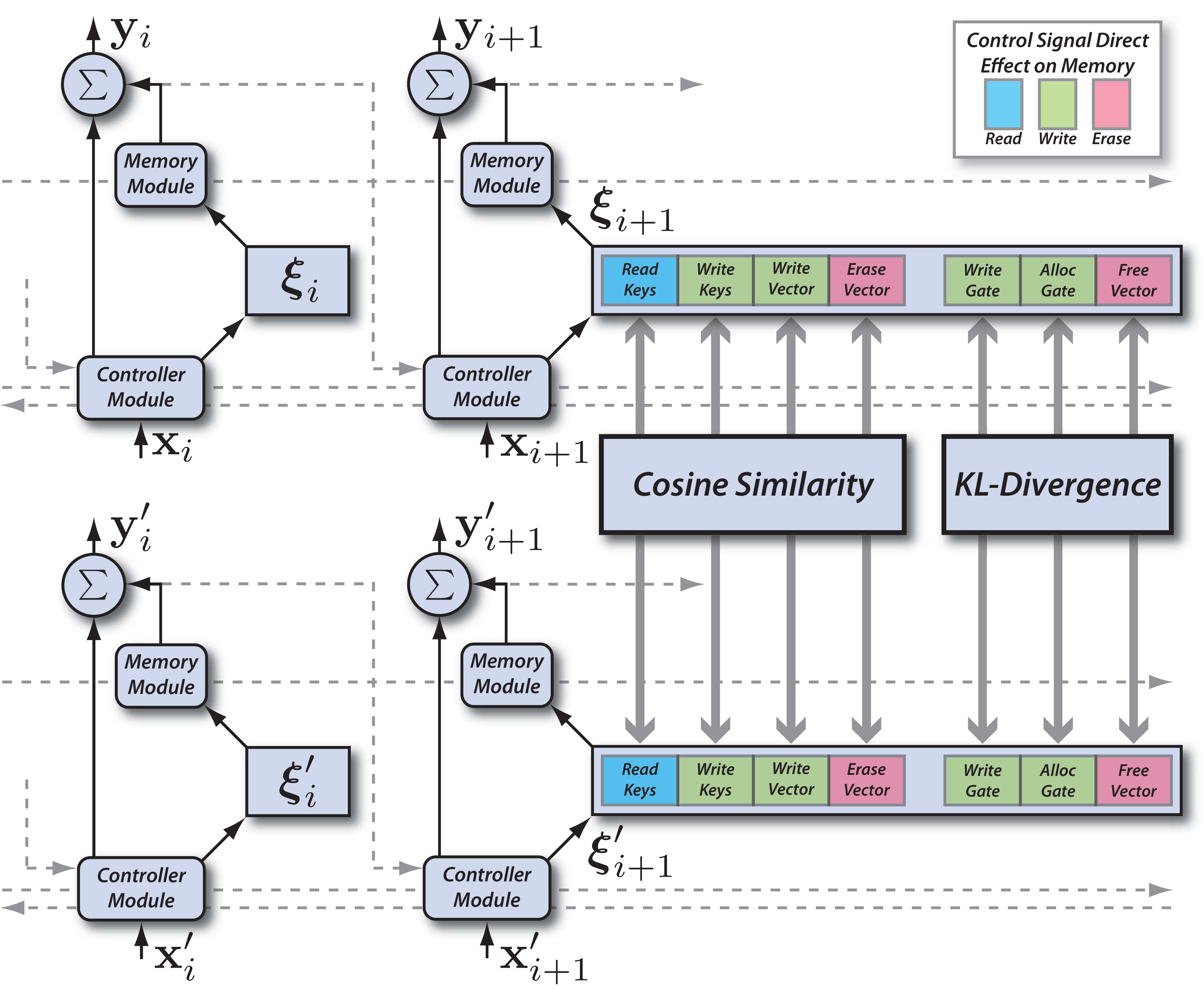}
\caption{(Top) Control signals from DNC with benign input can be compared with (Bottom) control signals from DNC with adversarial input.}
\label{fig:teaser}
\end{figure}
 
Differential neural computer (DNC) \cite{graves2016hybrid} was recently proposed as an extension of neural Turing machine (NTM) \cite{NTM}, with the addition of memory attention mechanisms that control where data is stored in the memory and temporal attention that records the order of events. Unlike the common feed-forward DNN and recurrent neural networks, DNC is Turing complete and adopts the Von Neumann architecture with a scalable memory module. DNC has achieved superior performance on several complex tasks that demand the representation of data structure and storage of data over long timescales, as demonstrated on bAbI question answering (QA) tasks, graph experiments (London Underground and family tree) and a block puzzle solving task \cite{graves2016hybrid}. 

Many adversarial attack studies have been conducted on feed-forward DNN and some on recurrent neural networks. Even with DNC's promise as a more universal approach to solving machine learning tasks, there lacks a in-depth study on whether such adversarial attacks also applied to it. Unlike common DNN, DNC separates its computational and memory capabilities into a DNN central controller and memory module respectively. This brings an opportunity to study how adversarial attacks affect the computational and memory components of a neural network separately. With the augmented memory enhancement, DNC tends to be more robust than a NTM. However, it is still unknown how the memory contributes to DNC's robustness against adversarial attacks.

Gradient based adversarial techniques, from computer vision domain, do not apply to our experiments due to discrete nature of input in the NLP domain. The replacement of a single word can drastically alter the semantics of a text or introduce grammatical errors. Previous NLP adversarial methods either erase or change words directly with domain-specific rules or require human intervention \cite{Alzantot,kuleshov2018adversarial,advcompre}.
To address these challenges, we propose scalable adversarial strategies that rely on metamorphic relations to generate attacks which are grammatically sound and preserve correct answers in QA tasks.
In this paper, we demonstrate that DNC can be vulnerable to our adversarial attacks on bAbI QA tasks, where it has originally achieved near-perfect performance \cite{rsDNC}. We also find that a larger size of the memory module provides limited benefit in DNC's robustness against such attacks. The effectiveness of such attacks are determined by their position, type of content and length, as shown by our experiments. Finally, we analyze the activities of DNC controller control signals and find that adversarial attacks disrupt the read, write and erase functions of the DNC.

The major contributions of this paper are the following:
(1) \textbf{First adversarial attack on DNC}, a neural network architecture that display state-of-the-art performance in various tasks, that demonstrates its vulnerabilities.
(2) \textbf{Pick-n-Plug and Pick-Permute-Plug}: Using two new automated and scalable strategies to generate grammatically correct adversarial attacks in the NLP QA domain.
(3) Evidence that larger memory size provides limited benefit in resisting such adversarial attacks. 
(4) Analysis of DNC's control signals, as illustrated in Figure~\ref{fig:teaser}, which shows that adversarial attacks disrupt the read, write and erase functions of the DNC. 

\section{Background and Related Work}
\subsubsection{Adversarial Attack}

Adversarial attacks were first discovered in computer vision domains \cite{szegedy2013intriguing}. Carefully perturbed images, with changes imperceptible to humans, can easily fool DNNs. Since then, we have witnessed an arms race between attackers \cite{iclr18-ae-boundary-analysis,iclr18-a-boundary-attack,iclr18-ae-natural} and defenders \cite{iclr18-b-counter,iclr18-b-defense-gan,iclr18-b-thermo}. The presence of adversarial examples have permeated into various computer vision tasks apart from visual classification, such as face recognition \cite{sharif2016accessorize}, object detection \cite{Xie_2017_ICCV}, semantic segmentation \cite{fischer2017adversarial}, generative modeling \cite{kos2018adversarial}, robustness testing \cite{felix_ase18_gauge,felix_arxiv18_fuzz}, \etc.

Although adversarial attack techniques are extensively studied in computer vision domain, 
there is limited work conducted in NLP domain \cite{advcompre,Alzantot,kuleshov2018adversarial,advrnnpapernot}.
One challenge lies in the discrete nature of word inputs in NLP which makes implementation of gradient-based perturbation methods challenging. Different from adversarial attack for images where small pixel changes are very unlikely to alter the correct class of an image, a change of word in a body of text may completely change the meaning of it under a particular NLP task or introduce grammatical errors. 

For adversarial attacks in NLP, Jia \cite{advcompre} proposed to add distracting sentences to the original text, which are generated from the task's question with rules. However, the grammatical correctness and the preservation of correct answer rely on manual checks, which makes such method difficult to scale.
Word substitution based attacks \cite{Alzantot,papernot2016crafting,kuleshov2018adversarial} were also proposed, by changing words in original text with synonyms.
In \cite{Alzantot}, a portion of the substituted text are interpreted by humans to be a different class, highlighting the challenge of creating adversaries that avoid changing the meaning of original text.
In this paper, we attempt to overcome these challenges with adversaries generated based on metamorphic relations. Our adversarial methods are scalable and do not disrupt the information from the original text for QA tasks.

\subsubsection{Differential Neural Computer (DNC)}
A DNC is a DNN augmented with an external memory module in the form of a matrix $M \in \mathbb{R^{\mathrm{N \times W}}}$ \cite{graves2016hybrid}. The DNN of DNC acts as the controller module, whose operations can be learn with gradient descent, while the external memory matrix serves as a module for data storage. The DNC's controller and memory module are like its CPU and RAM respectively. The DNC's memory can be written to and accessed by the controller at each input time-step.

At a particular time-step, the DNC controller takes in an input vector $\mathbf{x}_t \in \mathbb{R^{\mathrm{X}}}$,  and a set of read vectors from the previous time-step $\boldsymbol{\mu}_{t-1} \in \mathbb{R^{\mathrm{P}}}$.
\begin{align}
(\mathbf{v}_t, \boldsymbol{\xi}_t)=\mathrm{Controller}([\mathbf{x}_t, \boldsymbol{\mu}_{t-1}], \theta_c)
\end{align}
$\theta_c$ is the controller's trainable weight parameters, $\mathbf{v}_t \in \mathbb{R^{\mathrm{C}}}$ is the controller output while $\xi_t$ is a set of control signals.

The controller uses its write and read heads in order to manage the memory matrix. At each time-step, the controller's set of control signals $\boldsymbol{\xi}_t$ represents the operations of these heads. These control signals can be categorized into gates, keys or vectors. Their values determine how, where and what is being read, written and erased from the memory matrix. 
A series of operations in the memory module with the control signals and its current memory matrix $M_t$ erase and write new data and produces a concatenation of read vectors $\boldsymbol{\mu}_t \in \mathbb{R^{\mathrm{P}}}$.
\begin{align}
\boldsymbol{\mu}_t = \mathrm{MemoryModule}(\boldsymbol{\xi}_t, M_t)
\end{align}

The final output of the DNC is a sum of weighted controller output and weighted concatenation of read vectors from the memory module.
\begin{align}
\mathbf{y}_t=W_v \mathbf{v}_t + W_{\mu} \boldsymbol{\mu}_t + \mathbf{b}_t
\end{align}
where $W_v \in \mathbb{R^{\mathrm{Y \times C}}}$, $W_{\mu} \in \mathbb{R^{\mathrm{Y \times P}}}$ and $\mathbf{b}_t \in \mathbb{R^{\mathrm{Y}}}$ are trainable weights of the DNC.

There are two mechanisms for writing into memory $M_t$: least used location and content-based addressing. Least used location mechanism find least utilized memory locations for new information to be written in. Content-based addressing find memory locations which have the highest similarity with the controller's write keys. Allocation gate determines how much of these two mechanisms to use in order to write new information.

For reading, there are also two mechanisms in the original DNC paper. The first is content-based addressing which is very similar to that of the writing operation. Read keys, rather than write keys, are used to find locations to read from. The second mechanism, called temporal linkage mechanism, helps retrieve information in chronological order of when they are written. In an improved version of DNC, this mechanism is dropped for bAbI tasks to increase its memory efficiency \cite{rsDNC}. Free gate values are used to forget data which were recently accessed by the controller read heads.

\subsubsection{bAbI Dataset}
DNC that we are evaluating has shown near-perfect performance on the bAbI dataset which makes it a good candidate to evaluate the DNC's vulnerability to attacks. The bAbI dataset has 20 question \& answer tasks of different themes to evaluate a range of logical reasoning capabilities. Each task contains stories where each story has one or more following questions with answers that can be derived from the story.
After removing digits from the stories, all words are converted into lower case and splitted into word tokens. The whole vocabulary contains 156 unique words and three symbols: `?', `.', `-'. The `-' is used to indicate positions in the stories where answers are required.
The performance of bAbI tasks is evaluated by word error rate (WER) which is the rate of incorrect answers over total number of answers.

\section{Metamorphic Relation-Based Adversaries}\label{sec:method}
In a general classification task, a successful adversarial attack is a modified input $x'$ of an original input $x$ such that it causes the originally correct prediction $y = f(x)$ from a classifer $f$ to be incorrect such that $y' = f(x')$ and $y' \neq y$.
This definition can be extend to NLP question answering (QA) tasks such that $x$ and $x'$ can be generalized to $X$ and $X'$, where $X = [\mathbf{x}_1, \mathbf{x}_2, \ldots, \mathbf{x}_m]$,  $X' = [\mathbf{x}'_1, \mathbf{x}'_2, \ldots, \mathbf{x}'_{m'}]$ and $\mathbf{x}_i, \mathbf{x}_i' \in \mathbb{R^{\mathrm{X}}}$ are one-hot vectors of words in the original and adversarial input sequence respectively. $y$ can also be extended to become $Y$, where $Y=[\mathbf{y}_1,\ldots]$ which is a sequence of correct answer one-hot word vectors $\mathbf{y}_i$. 

\subsubsection{Metamorphic Transformation}
To generate adversarial attacks which do not change the original answer to bAbI QA tasks, we draw inspiration from metamorphic relations. An example of metamorphic relations for sine function is $\sin(x + \pi) = \sin(x)$. Metamorphic relations have been used in testing software \cite{chen1998metamorphic} and supervised classifiers \cite{xie2011testing}. Here, we define a metamorphic transformation $T$ as a function that maps an input $x$ to $x'$ which satisfy a metamorphic relation with $f$. More formally,
\begin{align}
x' = T(x)~~~~~~~~~~~~f(x) = f(x') \nonumber
\end{align}
$x$ is the original input and $x'$ is the output of a metamorphic transformation of $x$.

In the example of {\it sine} function where $f(x) = \sin{(x)}$, a valid metamorphic transformation is $T(x) = x + \pi$.
Similarly, in the QA tasks, the input $x$ and $x'$ can be generalized to $X$ and $X'$, \ie, $X' = T(X)$, $f(X) = f(X')$, where $f$ is an oracle that is \emph{always correct}, $X = [\mathbf{x}_1, \mathbf{x}_2, ..., \mathbf{x}_m]$,  $X' = [\mathbf{x}'_1, \mathbf{x}'_2, ..., \mathbf{x}'_{m'}]$ and $\mathbf{x}_i, \mathbf{x}_i' \in \mathbb{R^{\mathrm{X}}}$ are one-hot vectors of words in the original and transformed adversarial input sequence respectively. For any input $X'$ generated by the metamorphic transformation on $X$, the answer $f(X')$ would remain unchanged from $f(X)$ under the oracle $f$.

Consider a DNN model for a QA task as $f'$ where its prediction of an input sequence $X$ is $f'(X)$. If an adversarial input $X'=T(X)$ is generated with a metamorphic transformation $T$ (\ie, $f(X) = f(X')$) such that $f'(X) \neq f'(X')$, this would be a successful adversarial attack. 
\subsubsection{Pick-n-Plug}

\begin{figure}
  \begin{minipage}[c]{0.49\columnwidth}
    \centering
    \includegraphics[width=0.8\columnwidth]{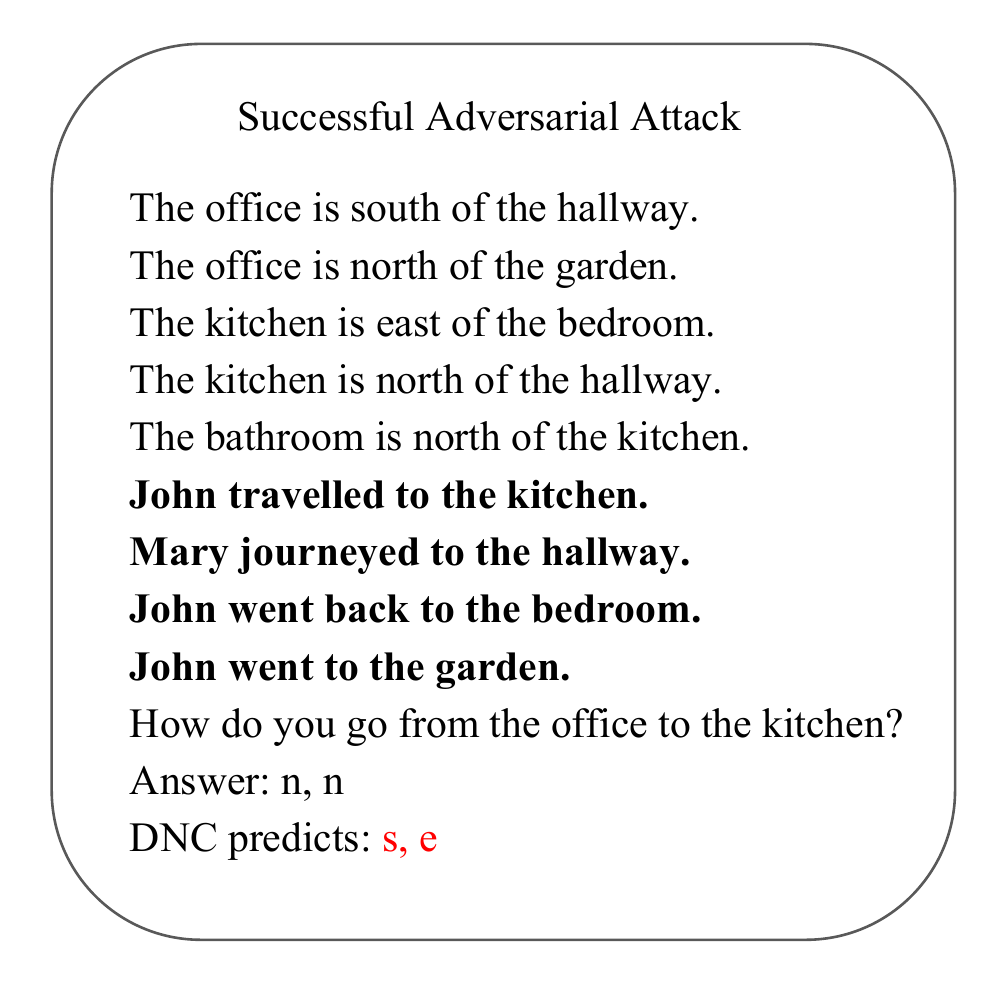}\\~\\
  \end{minipage}\hfill
  \begin{minipage}[c]{0.51\columnwidth}
    \caption{A successful Pick-n-Plug attack on target task \#19 with source task \#3, where the adversarial sentences (in bold) are picked from task \#3 and injected right before the question, after the story. In the presence of these adversarial sentences the DNC predicts `s', `e' which stands for `South' and then `East'. The DNC correctly predicts `n', `n' (`North' then `North') without adversarial sentences.}\label{fig:adv_example}
  \end{minipage}
  \vspace{-10mm}
\end{figure}

A metamorphic transformation can be composed of a series of $n$ operator functions $g_1, g_2,\ldots,g_n$, such that: $T(X) = g_n(\ldots g_2(g_1(X))\ldots)$.
We propose Pick-n-Plug which relies on metamorphic transformation $T_{\mathrm{pick-n-plug}}$ to generate adversarial attacks. It consists of a pick operator $g_{\mathrm{pick}}$ to draw adversarial sentences from a particular task (source task) and plug operator $g_{\mathrm{plug}}$ to inject these sentences into a story from another task (target task), without changing its correct answers. The $g_{\mathrm{pick}}$ step to draw sentences can be a random search, as we have demonstrated in our experiments, or other search methods. Pick-n-Plug requires no human intervention in generating adversarial attacks while ensuring grammatical correctness. Figure~\ref{fig:adv_example} shows an example of successful Pick-n-Plug attack where task \#19 and \#3 are the target and source task respectively. More formally, 
\begin{gather}
T_{\mathrm{pick-n-plug}}(X) = g_{\mathrm{plug}}(g_{\mathrm{pick}}(X)) = X' \mathrm{~~where} \nonumber\\
g_{\mathrm{pick}}(X) = (X, [S_1, \ldots, S_l]) \nonumber\\
g_{\mathrm{plug}}(X, [S_1, \ldots, S_l]) = X' \nonumber
\end{gather}
where $S_i = [\mathbf{x}^{S_i}_1, \mathbf{x}^{S_i}_2, \ldots, \mathbf{x}^{S_i}_j]$ is a sequence of word vectors in an adversarial sentence, $l$ is the number of adversarial sentences picked from the source task. $X'$ is an adversarial input as the final output of the Pick-n-Plug metamorphic transformation $T_{\mathrm{pick-n-plug}}$.
In our adversarial attacks, we identify a pair of a target task and a source task. One or more sentences $S_1, \ldots, S_l$ from the source task are picked, with an operator $g_{\mathrm{pick}}$ and then plugged $g_{\mathrm{plug}}$ into a story from the target task, with the aim of changing the DNC's answer from correct to incorrect. To maintain coherence within the injected text, blocks of consecutive sentences from the source task are used to attack target stories, with the first sentence in the block being randomly sampled.

$T_{\mathrm{pick-n-plug}}$ relies on a metamorphic relation between sentences and questions from several pairs of tasks such as task \#19 and \#3. Sentences from one task do not interfere with the information expressed by sentences from another task, and hence do not change the correct answer to their questions. For example, the directional information contained in Sentence 1: `\textit{The office is south of the hallway.}' from task \#19 is still preserved even when Sentence 2: `\textit{Mary journeyed to the hallway.}' from task \#3 is in the same story. To illustrate, for a story containing Sentence 1, if the correct answer to the question: `\textit{How do you go from the office to the hallway?}' is `\textit{north}', adding Sentence 2 `\textit{Mary journeyed to the hallway.}' to the story should not change the correct answer. Pairs of unrelated text corpora would potentially have this property.

While using the Pick-n-Plug to attack a target task, the choice of source task, number of injected sentences, and position of the adversarial injection can be varied. In the following sections of this paper, we show the effect of these factors on the DNC's performance.

\subsubsection{Pick-Permute-Plug}

We also propose Pick-Permute-Plug with metamorphic transformation $T_{\mathrm{pick-permute-plug}}$ to extend the adversarial capability of Pick-n-Plug. In Pick-n-Plug, the diversity of adversarial injected sentences is restricted by the text of the source task. With an additional $g_{\mathrm{permute}}$ operator after picking sentences ($g_{\mathrm{pick}}$) from a source task, words in a particular adversarial sentence can be permuted with `synonyms' to generate much wider range of possible attacks. Since only words in the in adversarial sentences are permuted, the correct answer is still preserved. $g_{\mathrm{permute}}$ can optionally rely on the original input $X$ to influence $S'_1, .., S'_l$ according to conditions such as to enforce similar grammatical tense or style of writing.
The Pick-Permute-Plug metamorphic transformation can be summarized as:
\begin{gather}
T_{\mathrm{pick-permute-plug}}(X) = g_{\mathrm{plug}}(g_{\mathrm{permute}}(g_{\mathrm{pick}}(X))) = X'\nonumber\\
\mathrm{where~~~~}g_{\mathrm{pick}}(X) = (X, [S_1, \ldots, S_l])\nonumber\\
g_{\mathrm{permute}}(X, [S_1, \ldots, S_l]) = (X, [S'_1, \ldots, S'_l])\nonumber\\
g_{\mathrm{plug}}(X, [S'_1, \ldots, S'_l]) = X'\nonumber
\end{gather}
and $S'_i = [\mathbf{x}^{S'_i}_1, \mathbf{x}^{S'_i}_2, \ldots, \mathbf{x}^{S'_i}_j]$ is an adversarial sentence from the ${permute}$ step such that $S'_i \neq S_i$ if one or more of its words have been substituted with synonyms.

In the same example above, the word `\textit{hallway}' in Sentence 2: `\textit{Mary journeyed to the hallway.}' can be substituted with `\textit{office}' under the $g_{\mathrm{permute}}$ operator to generate Sentence 3: `\textit{Mary journeyed to the office.}', before injecting into the story. For a story containing Sentence 1, if the correct answer to the question: `\textit{How do you go from the office to the hallway?}' is `\textit{north}', adding Sentence 3 to the story should also not change the correct answer.
Other words can also be selected to be permuted such as substituting the name word '\textit{Mary}' in Sentence 3 with '\textit{John}'.
The added flexibility allows for more control of the target DNC's predictions and behaviors with wider range of possible changes in the input sequence.
In practice, the $permute$ step could be executed by greedily permuting synonyms over the Pick-Permute-Plug process iteratively with respect to the DNC's output confidence, in order to induce prediction of a target output with high confidence.

\section{Experiments}\label{sec:exp}

\subsection{DNC Robustness against Adversarial Framework}

\subsubsection{Experiment Setup}
We conduct in-depth evaluation on the robustness of DNC using our adversarial framework. The DNC was jointly trained on en-10k data subset of all 20 bAbI tasks and obtained near-perfect performance in all the tasks. We conduct the following evaluations on the DNC model that was pretrained on all tasks and publicly released in \cite{rsDNC}.
In our adversarial attacks, we select 4 representative tasks (\#3, \#15, \#16 and \#19) of bAbI tasks to form 12 target-source task pairs and evaluate the DNC’s robustness on them.

\begin{table}[!htbp]
\centering
\caption{Word error rate (\%) of DNC on task \# 3, 15, 16 and 19 under Plug-n-Pick attacks. Test error rate without adversarial sentences for task \# 3, 15, 16 and 19 are 1.6\%, 0\% , 0\% and 0\%, respectively. `Full Block' adversarial injection means that a complete sequence of sentences from a sampled source task story is injected as adversarial sentences.}

\begin{tabular}{cc|cccc|c}
\toprule
\ \multirow{2}{*}{Source} & \multirow{2}{*}{Position} & \multicolumn{4}{c|}{\# of Adversarial Sentences} & \multirow{2}{*}{Full Block} \\
\cline{3-6}
& & 1 & 2 & 3 & 4 &\\\hline
\multicolumn{7}{c}{\emph{Under Adversarial Attacks on Task 3}} \\
\multirow{2}{*}{\#15} & before\_story & 3.6 & 2.8 & 3.3 & 4.8 & 21.0\\
& before\_question & 2.7 & 2.9 & 3.6 & 5.9 & 22.4\\
\multirow{2}{*}{\#16} & before\_story & 4.6 & 3.9 & 4.9 & 8.1 & 23.9\\
& before\_question & 2.5 & 2.6 & 4.4 & 6.3 & 26\\
\multirow{2}{*}{\#19} & before\_story & 4.5 & 6.7 & 8.3 & 9.5 & 12.0\\
& before\_question & \textbf{6.4} & \textbf{10.5} & \textbf{15.2} & \textbf{20.4} & \textbf{26.5}\\
\bottomrule
\multicolumn{7}{c}{\emph{Under Adversarial Attacks on Task 15}} \\
\multirow{2}{*}{\#3} & before\_story & 0.6 & 5 & 9.6 & 13.6 & 88.4\\
& before\_question & 0.7 & 3.9 & 8.4 & 13.9 & \textbf{98.5}\\
\multirow{2}{*}{\#16} & before\_story & 1.9 & 6.6 & 13.3 & 24.2 & 56.4\\
& before\_question & 1.4 & 5.4 & 11.9 & 23.1 & 60.3\\
\multirow{2}{*}{\#19} & before\_story & \textbf{4.4} & \textbf{22.4} & \textbf{40.9} & 53.7 & 64.2\\
& before\_question & 3.3 & 17.8 & 37.5 & \textbf{55} & 65.8\\
\bottomrule
\multicolumn{7}{c}{\emph{Under Adversarial Attacks on Task 16}} \\
\multirow{2}{*}{\#3} & before\_story & 0.4 & 0.9 & 0.7 & 0.7 & 24.5\\
& before\_question & 0.1 & 1.1 & 1.3 & 1.3 & \textbf{94.2}\\
\multirow{2}{*}{\#15} & before\_story & \textbf{2.8} & 2 & 3.9 & \textbf{7.7} & 27.4\\
& before\_question & 0.2 & 1 & 1.8 & 2.5 & 12.9\\
\multirow{2}{*}{\#19} & before\_story & 1.3 & 1.9 & 2.6 & 4.3 & 5.4\\
& before\_question & 0.6 & \textbf{2.5} & \textbf{4.1} & 6.9 & 10.8\\
\bottomrule
\multicolumn{7}{c}{\emph{Under Adversarial Attacks on Task 19}} \\
\multirow{2}{*}{\#3} & before\_story & 0.2 & 0.3 & 0.5 & 0.85 & 10.2\\
& before\_question & 0.3 & 0.5 & 1.05 & 1.75 & \textbf{51.4}\\
\multirow{2}{*}{\#15} & before\_story & 0.55 & 1.2 & 4 & 7.15 & 28.6\\
& before\_question & \textbf{0.9} & \textbf{2.95} & \textbf{8.55} & \textbf{14.8} & 37.0\\
\multirow{2}{*}{\#16} & before\_story & 0.4 & 2.4 & 4.45 & 6.75 & 19.1\\
& before\_question & 0.6 & 2.85 & 5.5 & 9.1 & 36.1\\
\bottomrule
\end{tabular}
\label{tab:t3_t15_t16_t19_adv}

\end{table}

The pick step is implemented as a random search for non-question sentences. To investigate the upper limit of the Pick-n-Plug adversarial effect, we also consider a scenario where bodies of text from the source task are picked from a story's start to the right before its first question, with no limit on the number of sentences.


\subsubsection{Results \& Discussion}

\subsubsection{Injection Position}
For both target tasks and all source tasks, when the number of adversarial sentences are larger, the adversarial sentences inserted right before questions have created stronger attack than sentences inserted at the start of the story, shown in Table~\ref{tab:t3_t15_t16_t19_adv}. For some cases where the number of adversarial sentences are smaller, the adversarial effect at the position before the story is larger. 

Intuitively, the effect of adversarial sentences inserted at the beginning of the story can be thought as to kick-start the DNC in the wrong direction to focus on details of the story that are not relevant in correctly answering the question. In contrast, the adversarial sentences right before the question might cause the DNC to erase data in its memory that is important to correctly answer the question at the end, as a price of storing data from the relatively more recent adversarial sentences. This implies that, as the length of adversarial sentences increases, the effect of the adversarial sentences in overwriting relevant data outweighs the effect of starting off the DNC's attention to the less relevant direction.

\subsubsection{Source Task}
Adversarial sentences from source task \#19 generally degrade the DNC's performance more than adversarial sentences from other source task (\#3, 15 and 16). A possible explanation may be that the distracting strength of adversarial sentences lies in the amount of information they carry. In task \#19, directional relationships between two location are expressed in each sentence. This translates to change to two entities’ attribute per sentence. For example, a sentence from task \#19 like ``\textit{The office is east of the garden.}'' encodes that for the `\textit{office}' entity's attribute that is east of `\textit{garden}’ and the `garden' entity's attribute that it is west of `\textit{office}'.

In task \#3, \#15 and \#16, only one attribute of a single object such as its location (\#3) or its identity (\#15) is altered per sentence. This means that sentences from task \#19 carry almost twice the amount of information about entities than sentences from \#3, \#15 and \#16, explaining the stronger adversarial effect.
It implies that adversarial sentences, within a word limit, generally have a more potent effect if the amount of information encoded is maximized in these sentences.

\subsubsection{Number of Adversarial Sentences}
For target tasks and all source tasks, adversarial blocks with more sentences generally contribute to more powerful adversarial attack in causing the DNC to predict incorrect answers. It can be interpreted intuitively that the adversarial effect of DNC storing irrelevant data and overwriting important information gets more pronounced with increasing amount of adversarial information.

In the case where the number of sentences is limited to 4, the DNC's performance in task \#15 degrade from $0$\% to $55$\% when attacked with task \#19 adversarial sentences injected right before the question. In the case where blocks of contiguous sentences from source task stories are injected without sentence limit, the DNC degrades from a perfect test performance of $0$\% error to $98.5$\% in target task \#15.

\subsection{Role of DNC Memory Module in Robustness}
The memory module in DNC contributes its state-of-the-art performance in bAbI tasks, but is still unknown what its role is in resisting adversarial attacks. Since one possible effect of adversarial sentences is to overwrite relevant stored information, a bigger size of the DNC's memory module might mitigate this effect by having more free space to write new information rather than overwriting these important data.

In the DNC architecture, its controller is meant to work with memory module of different sizes as long as the dimensionality of each memory row is the same. Since the learned parameters of DNC's controller are compatible with any number of rows the memory module can have, we use the same DNC's controller for all memory sizes.

We carry out adversarial attacks with Pick-n-Plug on DNC of memory size 0.5x, 0.75x, 1x, 2x, 4x, 8x, 16x and 32x of the original 192 memory rows. In one of the attacks, blocks of 4 adversarial sentences are injected before the question of the target task. In a stronger attack, one separate block of 4 adversarial sentences is injected before the question and another before the start of the target story. We carry out these attacks on target task \#3 using adversarial source task \#19 and vice versa.

The performance of DNC with these memory sizes are also evaluated on clean test dataset.

\subsubsection{Results \& Discussion}
\begin{figure}
\centering
\includegraphics[width=\columnwidth]{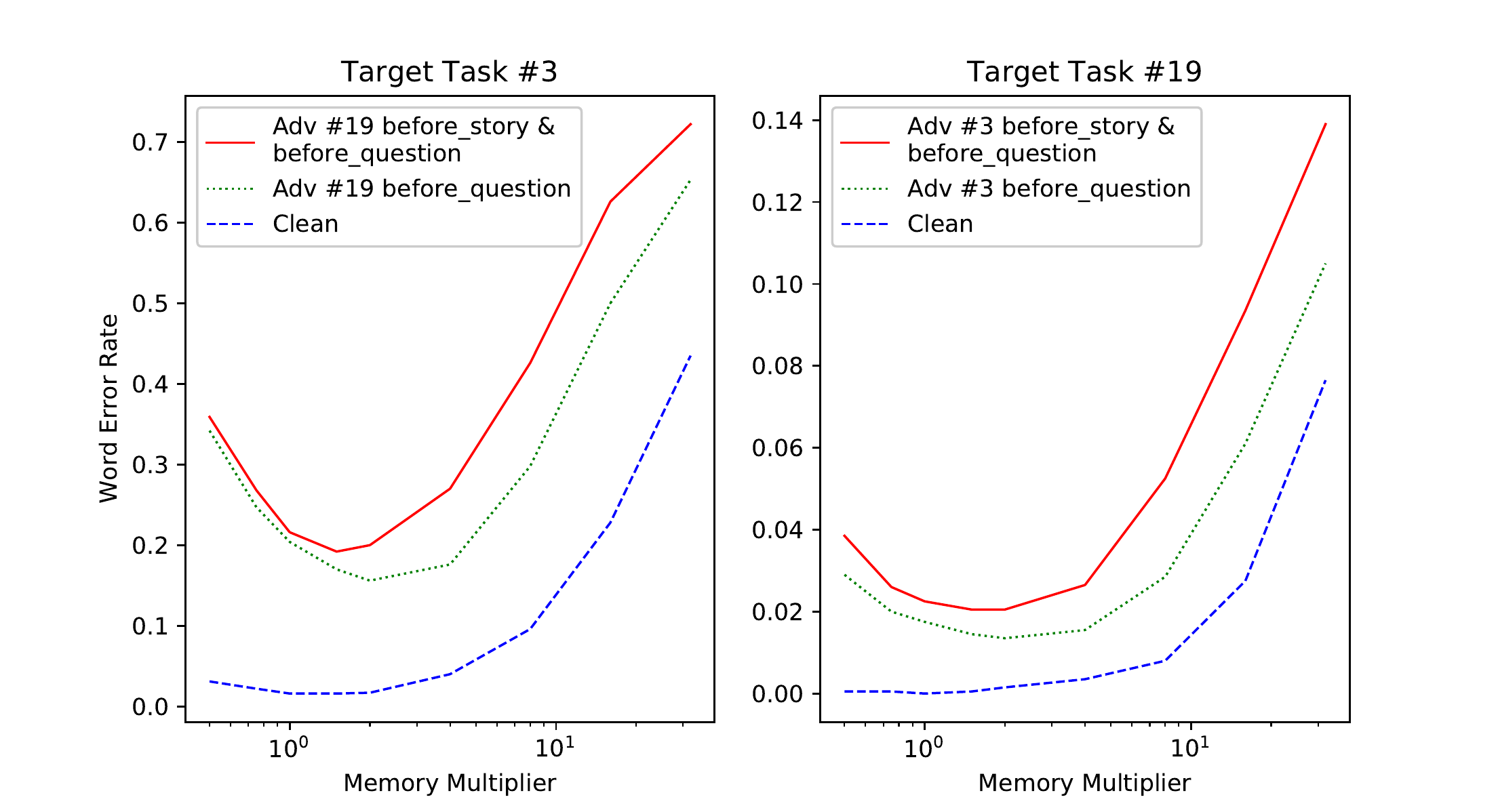}
\caption{(L) Word error rate (\%) of DNC with different sizes of augmented memory module on target task \#3 with adversarial source task \#19, and (R) on target task \#19 with adversarial source task \#3.}
\label{fig:memory}
\vspace{-4mm}
\end{figure}

Our evaluation results show that memory size plays a limited role in DNC's resistance against adversarial attacks from Pick-n-Plug, as shown in Figure~\ref{fig:memory}. The error rates of DNC drip to a minimum ($9$\% to $24$\% reduction in error from the original memory size) in our experiments when memory size is either 2x or 1.5x of its original. 

However, as the memory size increases further, the adversarial error rates increase even past the original error rate. This implies that there is still a significant robustness gap that larger DNC's memory size cannot close. The error rate of the DNC stands at $15.6$\% at its maximum robustness when task \#3 is attacked by block of 4 adversarial sentences injected before the question.

Its performance on clean test samples is also degraded, also shown in Figure~\ref{fig:memory}. This seems to occur earlier than the degradation of performance under adversarial attacks. An explanation for this might be that the DNC controller was not trained to optimally handle memory modules of sizes too far from its original size, even though they are compatible. This degradation due to unfamiliarity of DNC's controller with increasing memory size may outweigh the robustness benefit that larger memory size can offer, resulting in a maximum robust performance point close to the original memory size, 1.5 to 2x in our experiments.

\subsection{Adversarial Effect on DNC Controller Activities}

We investigate the behaviors of DNC controller under different inputs, 1) clean input (CE), 2) unsuccessful adversarial attack (UAA), and 3) successful adversarial attack (SAA),
by probing its control signals. The DNC controller outputs control signals to determine how, where and what content that is read, written and erased from the DNC memory module. These control signals comprise 3 gates, 2 keys and 2 vectors. Gate values are scalar that range from 0 to 1, while keys and vectors are W-dimensional vectors with real values.

The free gate value determines how much of recently accessed information to forget. Allocation gate shows how much the location of written content is influenced by availability of space in that location rather than the relevance with the information already stored in that location. The write gate determines how much to write into memory at a particular time step.

The write key and read key guide to memory locations with highest similarity to write new data and read stored information, respectively. Write vector and erase vector are the two vector values which determine what content to write into and erase from the memory, respectively.

We can probe how DNC controller's behaviors change by comparing how the control signals deviate under one input sequence to another. In our experiments, we compare the signals under different inputs with pairwise similarity metrics in such fashion: clean example with unsuccessful adversarial attack (CE-UAA), clean example with successful adversarial attack (CE-SAA), and unsuccessful adversarial attack with successful adversarial attack (UAA-SAA). The UAA is an example where the DNC's input sequence contains injected adversarial sentences and it still predicts the correct answer to the question. In SAA, the DNC predicts the incorrect answer.

For similarity comparison of 2 sequences of scalar values, like the gate values, we can use normalized KL-divergence. We can use cosine similarity to compare vector-based control signals like the write/read keys and write/erase vectors.

\begin{figure*}
\centering
\includegraphics[width=0.65\columnwidth]{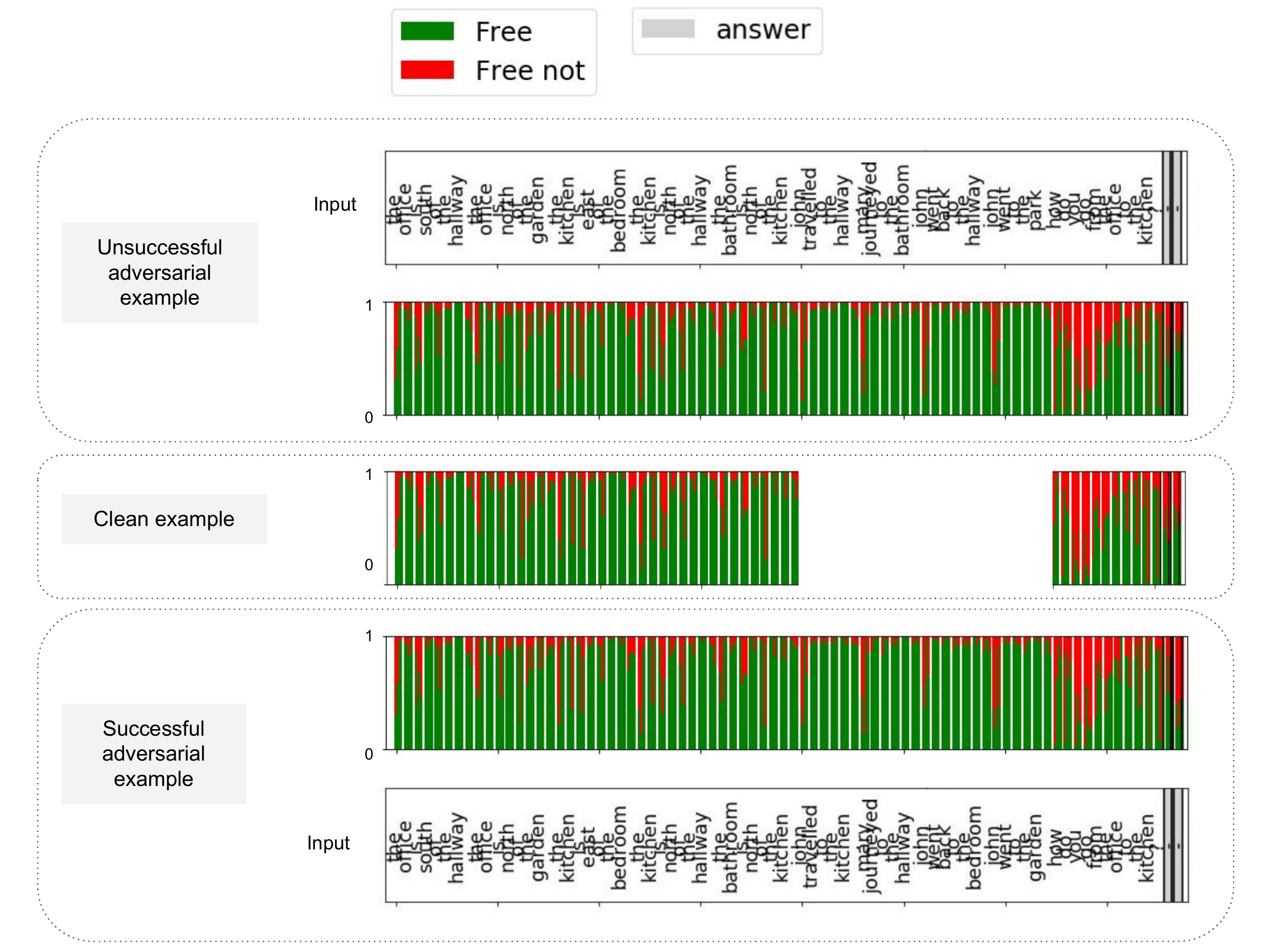}
\includegraphics[width=0.65\columnwidth]{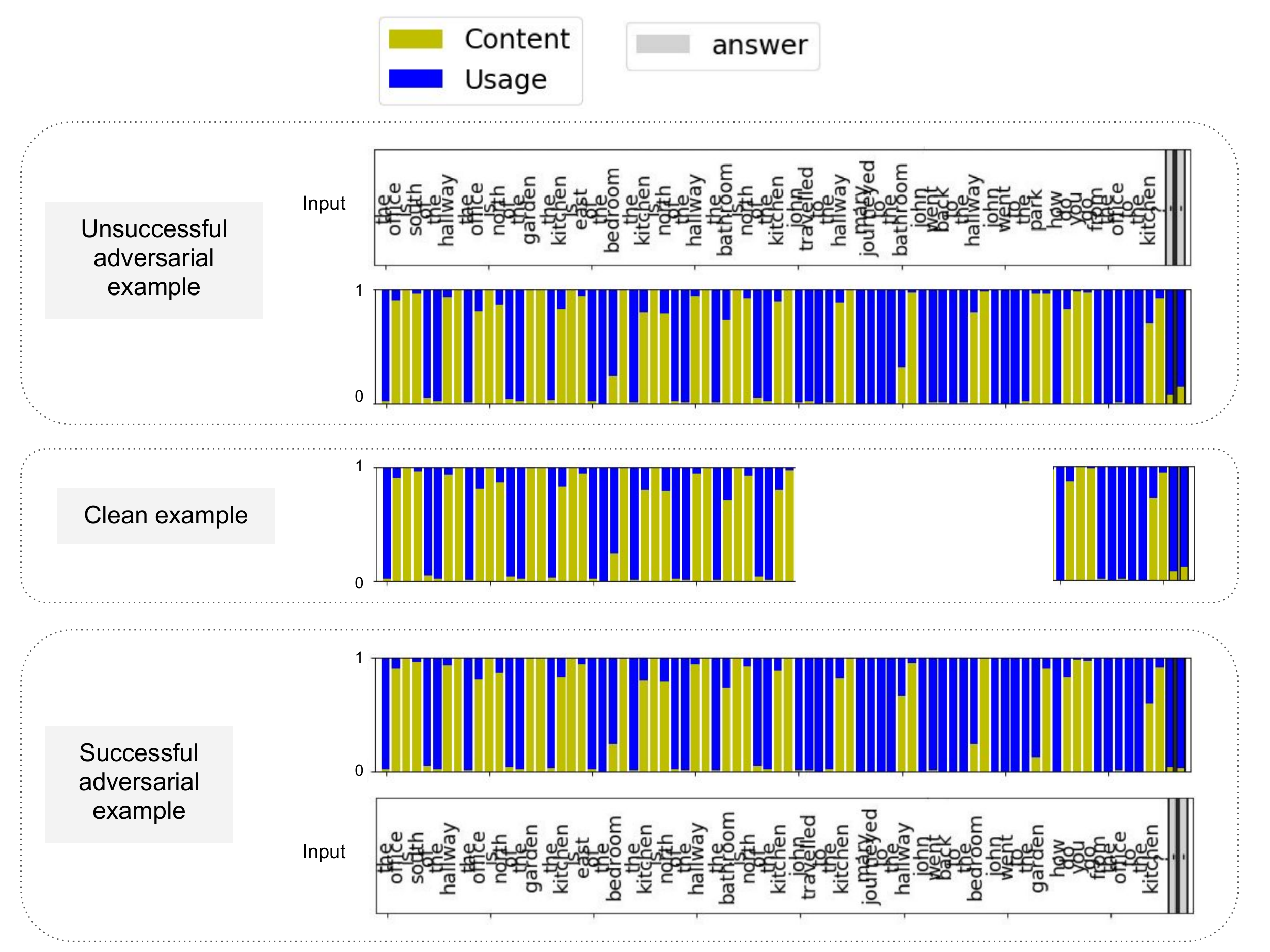}
\includegraphics[width=0.65\columnwidth]{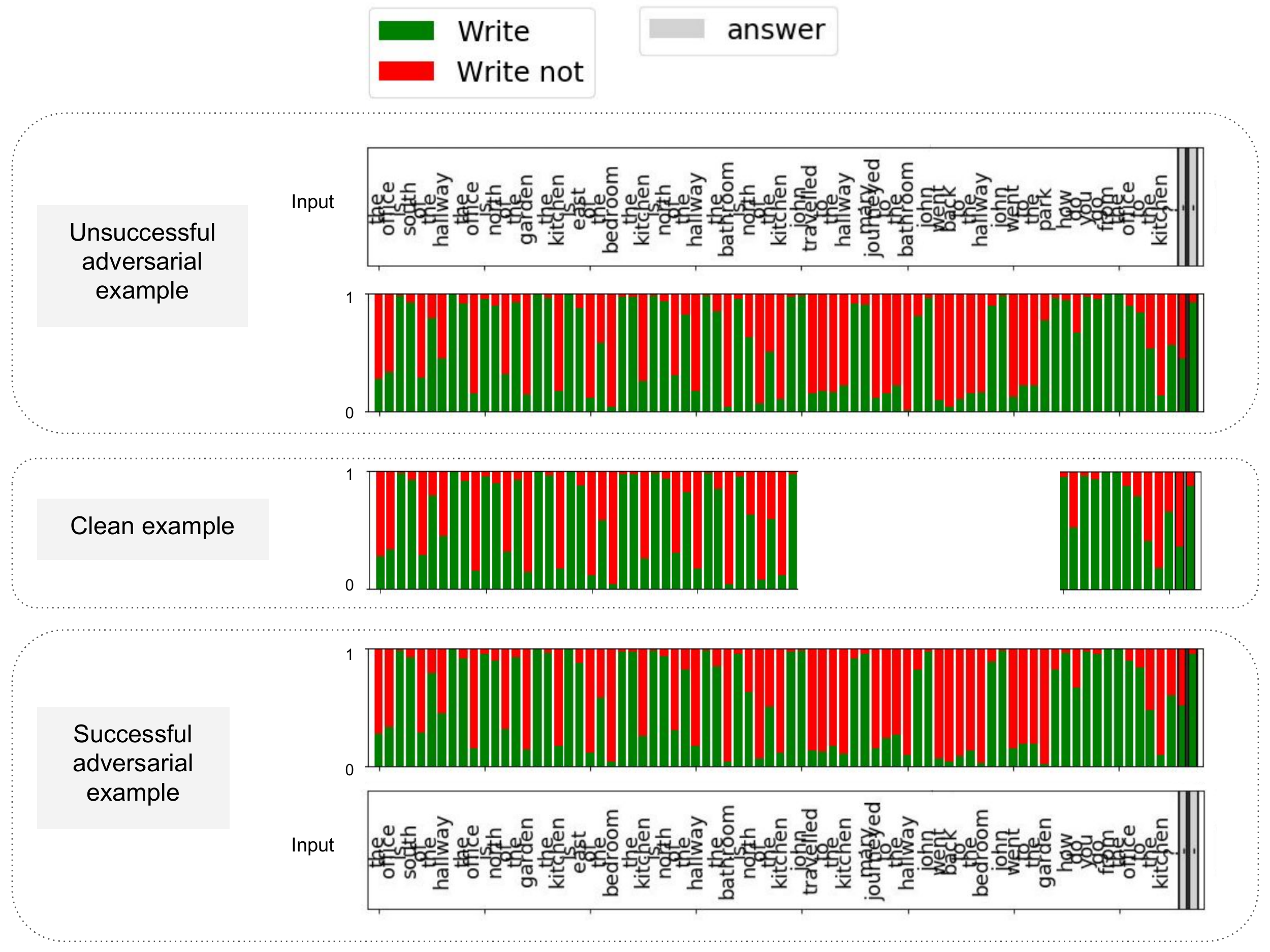}
\caption{(Top-Down) The free gate, allocation gate, and write gate values of DNC's read heads when the input sequence is (Row 1) an unsuccessful adversarial input, (Row 2) a clean example, and (Row 3) a successful adversarial input. }
\label{fig:combo}

\end{figure*} 

Every DNC input sequence in our experiments contain a story part and a question part. For an adversarial input sequences, we can further differentiate the story part into its clean and adversarial segments, to have a more fine grained picture of the control signals. We compared the DNC control signals under these separate parts of input sequence. 

Here, we describe how we sampled a CE, UAA and SAA that are closely related for a meaningful comparison. First, an UAA is sampled from a Pick-n-Plug run on target task \#19 that picks 4 adversarial sentences from source task \#3, and inject them right before every question. To generate closely related versions of this UAA sample, we conduct a run of Pick-Permute-Plug where 4 location words in this UAA's adversarial sentences are permuted with a set of 8 synonyms in a brute force manner, while the rest of the adversarial sentences remains the same. Among these new adversarial examples which successfully caused the DNC to predict incorrect answer, a SAA was sampled. The CE equivalent of the UAA and SAA is one with all 4 adversarial sentences removed. All words in the story and question segments of CE, UAA and SAA are the same. In the adversarial segments of UAA and SAA, up to 4 of the location words can differ while the other words and length of the word sequences are the same.

\subsubsection{Results \& Discussion}

For all pair-wise comparisons and for all of DNC's controller keys and vectors~(see Table~\ref{tab:cos_sim_combo}), the mean cosine similarities at the story segment are significantly higher than the mean cosine similarities at the question segment. This indicates that the main adversarial effect in disrupting the DNC's controller keys and vectors emerge mainly at sections after the adversarial sentences are injected rather than before that, despite DNC controller's bidirectional architecture. 

When compared with CE, the cosine similarities of controller keys and vectors from SAA is lower than that from UAA. This suggests that the keys and vectors from a clean example are disrupted more under a successful attack than an unsuccessful one.

\begin{figure}
\centering
\includegraphics[width=\columnwidth]{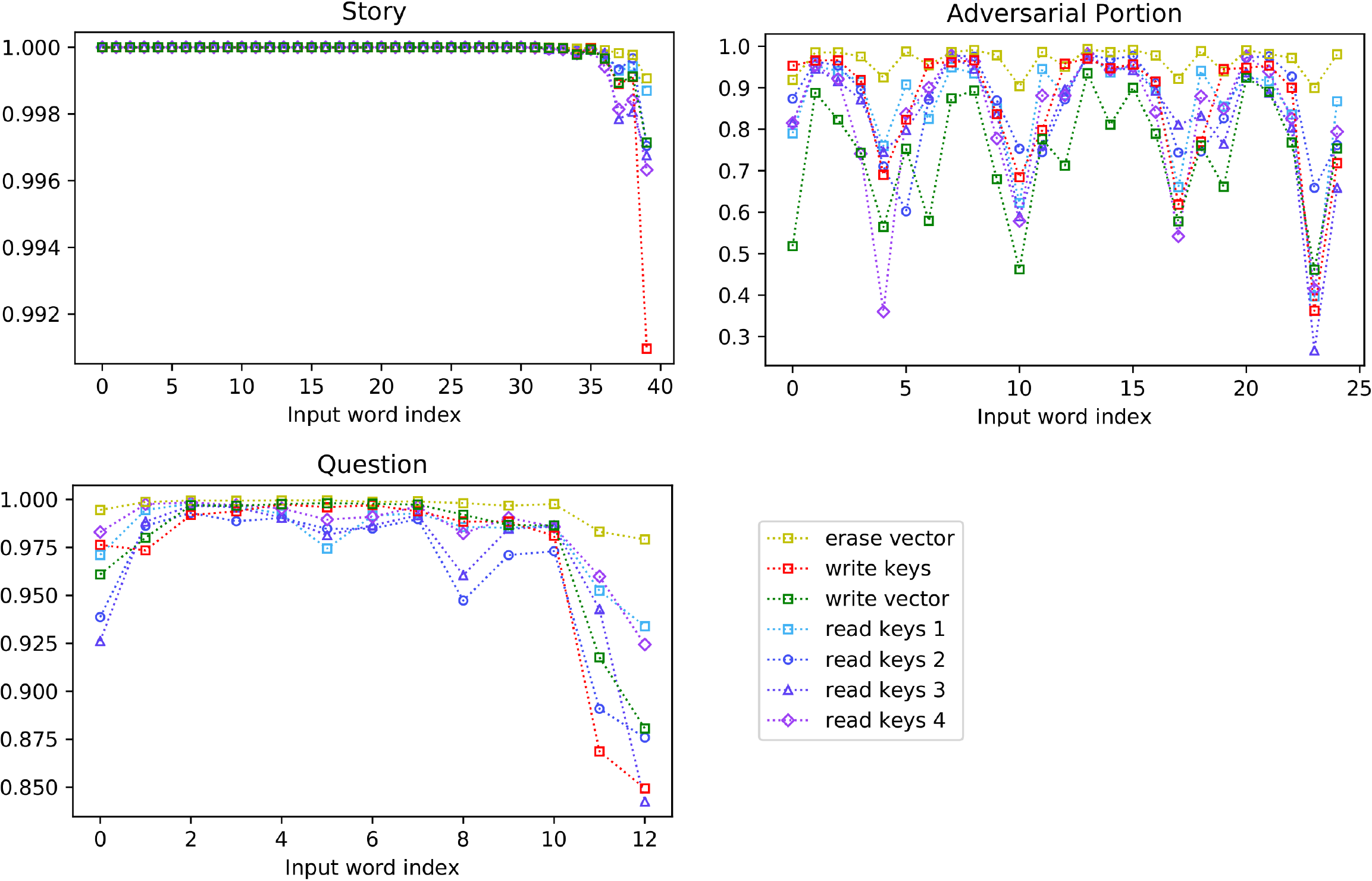}
\caption{Cosine similarity of DNC's keys and vectors, between when the DNC is presented with a unsuccessful adversarial attack and when it is presented with a successful adversarial attack. (Top left) The cosine similarity values at the story, (Top right) adversarial portion, and (Bottom) question of a sample QA from task \#19.}
\label{fig:cos_sim_UAA_SAA}

\end{figure}

When comparing the DNC's controller keys and vectors under UAA and SAA, the cosine similarities between all of them - the write keys, read keys, write vectors and erase vectors - are lowest in the adversarial segment. Upon closer look into the cosine similarities of these keys and vectors in the adversarial segment, we can see from Figure \ref{fig:cos_sim_UAA_SAA} that there are some sharp dips in the similarities of the keys and vectors. These dips are at the time-step where the input words are different. The four dips correspond to [`\textit{hallway}', `\textit{bathroom}', `\textit{hallway}', `\textit{park}'] in the UAA and [`\textit{kitchen}', `\textit{hallway}', `\textit{bedroom}', `\textit{garden}'] in the SAA. 

Looking at the cosine similarities at question segment, there is a sharp dip at the end where the answer is expected from the DNC, while the similarities are relatively stable at previous time steps. This suggests that the adversarial sentences has a latent effect on the keys and vectors which emerges in critical segments such as when information is retrieved to answer a question. This latent effect is very likely to be strengthened by the DNC memory module since it excels in storage of information.
These observations indicate that disruptions to these keys and vectors, which are involved in DNC's write, read and erase operations, play a part of the overall adversarial effect from a successful attack. For gate values, we find no obvious difference between the patterns of the gate values when DNC is presented with a CE, UAA and SAA from a general view (see Figure~\ref{fig:combo}).

When the KL-divergence is used to compare the gate values, significant patterns appear at the different segments of the input sequences. For all 3 type of gate values at the story segment~(see Table~\ref{tab:kl_combo}), the KL-divergence of all pair-wise comparisons are significantly lower than the KL-divergence at the question segment, with 2 to 3 order of magnitude difference. This implies that the main adversarial effect on DNC's 3 gate values emerge after the injection of adversarial sentences, rather than before that.

\begin{table}[!htbp]
\centering

\caption{Mean cosine similarity between DNC's keys and vectors as a comparison how different these control signals are with different inputs. The input can be a clean example (CE), unsuccessful adversarial attack (UAA) or successful advers. attack (SAA).}

\begin{tabular}{clccc}
\toprule
~ & ~ & Story & Adversary & Question \\ \cline{1-5}
\multirow{3}{*}{Write Keys} & CE-UAA & 0.9969 & - & 0.9790\\
~ & CE-SAA & 0.9962 & - & \textbf{0.9627}\\ \cline{2-5}
~ & UAA-SAA & 0.9997 & \textbf{0.8593} & 0.9689\\
\bottomrule
\multirow{3}{*}{Read Keys} & CE-UAA & 0.9981 & - & 0.9749\\
~ & CE-SAA & 0.9980 & - & \textbf{0.9652}\\ \cline{2-5}
~ & UAA-SAA & 0.9999 & \textbf{0.8417} & 0.9737\\
\bottomrule
\multirow{3}{*}{Write Vectors} & CE-UAA & 0.9988 & - & 0.9799\\
~ & CE-SAA & 0.9988 & - & \textbf{0.9653}\\ \cline{2-5}
~ & UAA-SAA & 0.9999 & \textbf{0.7400} & 0.9761\\
\bottomrule
\multirow{3}{*}{Erase Vectors} & CE-UAA & 0.9996 & - & 0.9975\\
~ & CE-SAA & 0.9995 & - & \textbf{0.9942}\\ \cline{2-5}
~ & UAA-SAA & 1.000 & \textbf{0.9660} & 0.9957\\
\bottomrule
\end{tabular}
\label{tab:cos_sim_combo}

\end{table}

\begin{table}[!htbp]
\centering

\caption{KL-divergence of DNC controller's gate values to compare how they changes with different inputs.The input can be a clean example (CE), unsuccessful adversarial attack (UAA) or successful adversarial attack (SAA). Free gate values from only 1 of the 4 DNC's read heads are compared here. The KL-divergences of free gate values from the other 3 read heads show similar spread of maximum values.}

\begin{tabular}{clccc}
\toprule
~ & ~ & Story & Adversary & Question \\\hline
\multirow{6}{*}{Free Gate} & CE-UAA & 3.735E-05 & - & 0.002472\\
~ & UAA-CE & 3.710E-05 & - & 0.002515\\
~ & CE-SAA & 2.526E-05 & - & \textbf{0.02041}\\
~ & SAA-CE & 2.529E-05 & - & 0.01566\\ \cline{2-5}
~ & UAA-SAA & 2.212E-06 & 0.005523 & \textbf{0.02470}\\
~ & SAA-UAA & 2.234E-06 & 0.006529 & 0.01791\\
\bottomrule
\multirow{6}{*}{Alloc Gate} & CE-UAA & 0.0003625 & - & 0.001112\\
~ & UAA-CE & 0.0003883 & - & 0.001184\\
~ & CE-SAA & 0.0003089 & - & \textbf{0.01512}\\
~ & SAA-CE & 0.0003319 & - & 0.01148\\ \cline{2-5}
~ & UAA-SAA & 3.730E-06 & \textbf{0.2279} & 0.02437\\
~ & SAA-UAA & 3.760E-06 & 0.1596 & 	0.01682\\
\bottomrule
\multirow{6}{*}{Write Gate} & CE-UAA & 0.0003325 & - & 0.005531\\
~ & UAA-CE & 0.0003171 & - & 0.005586\\
~ & CE-SAA & 0.0002664 & - & \textbf{0.007648}\\
~ & SAA-CE & 0.0002548 & - & 0.007347\\ \cline{2-5}
~ & UAA-SAA & 1.369E-05 & \textbf{0.2408} & 0.001681\\
~ & SAA-UAA & 1.389E-05 & 0.1101 & 0.001621\\
\bottomrule
\end{tabular}
\label{tab:kl_combo}
\end{table}
At the question segment, the KL-divergence of DNC's gate values from CE under SAA is higher than that under UAA for all the 3 gate types. This suggests that the gate values deviate more from a clean input under a successful attack than an unsuccessful one.
These two observations on the gate values, in the story and question segments, resonates with the above observations on the controller keys and vectors in these word segments.

When comparing the DNC's gate values under UAA and SAA, the KL-divergence of the allocation and write gate values in the adversarial segment is the largest. This might be due to the presence of 4 different location words in the adversarial sentences of UAA and SAA. In contrast, the KL-divergence of free gate values is larger in the question segment than in the story or adversarial segment, despite the fact that the words in UAA's and SAA's question segment are the same. This implies that the adversarial sentences have different effects on the gate control signals.
These observations suggest that disruptions to these gates, which are responsible for the DNC's erase and write operations, play a part of the overall adversarial effect from a successful attack.

\section{Conclusions}\label{sec:concl}
In this paper, we propose a
metamorphic relation based adversarial attack framework for DNCs, and show its attacks can cause multi-faceted disruptive effects on the read, write and erase functions of the DNC. Our in-depth evaluation on bAbI logical question answering task demonstrates that the current DNCs still face robustness issues, despite with larger memory size. We hope our framework motivates more work about adversarial attacks in NLP and more extensive studies on DNCs, towards constructing robust, automatically programmed, general purpose learning machines.

\bibliographystyle{IEEEtran}
\bibliography{reference}

\end{document}